\documentclass{svproc}
\usepackage{url}

\usepackage{amssymb}
\usepackage[hidelinks]{hyperref}
\usepackage{graphicx}

\begin{document}
\mainmatter              
\title{VitrAI - Applying Explainable AI in the Real World}
\titlerunning{VitrAI - Applying Explainable AI in the Real World}  
%
\author{Marc Hanussek\inst{1} \and Falko Kötter\inst{2} \and Maximilien Kintz\inst{2} \and Jens Drawehn\inst{2}}
\authorrunning{Hanussek et al.} 

\institute{University of Stuttgart IAT, Institute of Human Factors and Technology Management, Stuttgart, Germany,\\
\email{marc.hanussek@iat.uni-stuttgart.de},\\ WWW home page:
\texttt{https://www.iat.uni-stuttgart.de/en/}
\and
Fraunhofer IAO, Fraunhofer Institute for Industrial Engineering IAO,\\
Stuttgart, Germany}

\maketitle              

\begin{abstract}
With recent progress in the field of Explainable Artificial Intelligence (XAI) and increasing use in practice, the need for an evaluation of different XAI methods and their explanation quality in practical usage scenarios arises. For this purpose, we present \emph{VitrAI}, which is a web-based service with the goal of uniformly demonstrating four different XAI algorithms in the context of three real life scenarios and evaluating their performance and comprehensibility for humans. This work reveals practical obstacles when adopting XAI methods and gives qualitative estimates on how well different approaches perform in said scenarios.
\end{abstract}
\section{Introduction}
\label{sec:introduction}

The successful adoption of Artificial Intelligence (AI) relies heavily on how well decision makers can understand and trust it~\cite{doshivelez2017rigorous}. In the near future, AI will make far reaching decisions about humans, for example in self-driving cars, for loan applications or in the criminal justice system. Examples like COMPAS, a racially biased algorithm for parole decisions show the need for reviewing AI decisions~\cite{dressel2018compas}.

But complex AI models like Deep Neural Networks (DNNs) are hard to understand for humans and act as de-facto black boxes~\cite{mittelstadt2019explainingexplanations}. In comparison, simpler models like decision trees are more understandable for humans, but lack prediction accuracy~\cite{gunning2017explainable}, though some argue this is not necessarily true~\cite{rudin2019stopblackbox}. The field of Explainable Artificial Intelligence aims to create more explainable models while still achieving high predictive power. Current discussions in research suggest that explainability will lead to increased trust, realistic expectations and fairer decisions~\cite{doshivelez2017rigorous}. 

\emph{Post-Hoc explanation techniques} work by creating a simplified, interpretable explanation model that approximates the behavior of the black box~\cite{slack2020foolinglime}. However, this simplification is not without risk, as by definition the precise workings of the black box are not shown~\cite{slack2020foolinglime}. In contrast, \emph{interpretable models} aim to make the actual model used for predictions more transparent~\cite{rudin2019stopblackbox}. Basic examples of interpretable models are decisions trees and Bayesian rules lists~\cite{letham2015interpretable}. Neural networks can be made interpretable by tracking the degree to which each part of the input data contributed to the decision~\cite{zhang2018interpretablenn}.

Though the term \emph{model interpretability} is widely used in literature, it lacks an agreed-upon definition~\cite{lipton2018mythosinterpretability}. The same is true for the term \emph{explanation}, in particular what is a valid explanation and how to measure explanation quality. This stems from the difference between causality and correlation. In the aforementioned COMPAS example, it is unclear if race is used as a decision criterion, but still correlates strongly with the decision~\cite{dressel2018compas}\cite{rudin2019stopblackbox}. In another example, the model differentiated between pictures of huskies and wolves by the presence of snow, as wild animals tend to be photographed in the wilderness. To a human, this may be a bad explanation, though it correctly reveals the inner workings of the model~\cite{10.1145/2939672.2939778}. Therefore, it is necessary to distinguish the quality of explanations according to different purposes pursued.

For \emph{data scientists}, explanation quality hinges on the degree with which the actual decision criteria are shown. A high-quality explanation for wrong decisions would immediately show the human that the solution is low-quality, ideally giving an indication of a systematic error (e.g. due to bias in the input data). For \emph{end users}, explanation quality hinges on the degree to which an AI decision can be assessed based on the explanation. For example, if an AI identifies business cases from letters, the most relevant keywords could be highlighted so the human can decide if the decision is correct or if a manual reclassification is necessary. An explanation of the model's workings is secondary in comparison to a decision basis for human validation.

Ideally, AI reasons in the same way as humans, so that the explanation can be understood by humans. But this is not necessarily the case. There are many problems that are hard for AI and easy for humans (and vice versa). One example is general image classification, e.g. of animals. To humans differentiating between cats and dogs is trivial and an explanation provides little value to end users. Thus, evaluating explanations must also take into account tasks that are non-trivial for humans as well, for example identifying brands of vehicles.

For the use of AI in practice, it is necessary to provide high-quality explanations for end users. Thus, for developers of AI-based solutions it is necessary to evaluate explanation quality from an end-user point of view. Further research is necessary in what constitutes explanation quality in practical use and how that quality can be systematically evaluated, and furthermore measured.

In this work we present \emph{VitrAI}, a platform for demonstrating and evaluating XAI. We outline the main sections of the platform and XAI methods. We present a preliminary evaluation of explanation quality for theses tasks and describe how we plan to systematically evaluate explanation quality and compare human and AI explanations.

\section{Related Work}
\label{sec:relatedwork}

In this section, we investigate related work in the areas of (a) defining explainability and related terms and (b) benchmarking explainability.


\cite{lipton2018mythosinterpretability} defines desiderata for interpretability: Trust, causality, transferability, informativeness, fair and ethical decision making. Interpretability can be achieved by transparency or post-hoc explanations. Model properties that confer transparency are simulability, decomposability and algorithmic transparency. Results may be explained post-hoc by examples, verbal reasoning, visualizations, or local explanations. The author stresses the importance of a specific definition and goal when measuring interpretability and cautions against limiting the predictive power of AI for interpretability.

\cite{zhou2020intelligibility} argue that interpretability is not an intrinsic property of a model, but a property in perspective to a specific audience: Engineers, users or affectees, each of which have different goals and requirements regarding interpretability, and interpretability needs to be measured in regards to these goals.

\cite{montavon2018methods} detail current efforts to measure intelligibility, for example introspective self-reports from people, questionnaires and propose quantitative measurements for evaluating explanation quality, in particular continuity and selectivity.

In~\cite{das2020xaisurvey} an explanation is defined in the context of XAI as ``a way to verify the output decision made by an AI agent or algorithm'', which corresponds to the explanation quality for end users we have outlined. In addition, this work gives an overview of desired qualities promoted by explanations (trust, transparency, fairness) and explains different techniques for evaluating XAI algorithms. 


\cite{zhou2020intelligibility} outline possibilities for evaluating explainability, noting that human understanding of systems is implicit and that a benchmark would need to encompass enough experience for a human to build such an implicit mental model. They note however a lack of explicit measures of explainability. Interpretability is different from explanation accuracy, as falsehoods such as simplifications can aid overall understanding.

\cite{zhou2020assessingposthoc} propose leveraging existing work in the field of learning science by interpreting an AI's user as a learner and an explanation as learning content. They compare pairs of explanations in an educational context with user trials and semi-structured interviews. Using interview transcripts, they compared the user experience of the explanations. As postulated in~\cite{zhou2020intelligibility}, it was shown that helpfulness of different explanations depends on users' educational background.

\cite{geirhos2018humandnnbenchmark} investigates the capabilities of Deep Neural Networks to generalize by comparing DNN performance on distorted images with human performance. The results show that DNNs decline rapidly when images are distorted compared to humans, unless DNNs were trained for a specific kind of distortion.

\cite{slack2020foolinglime} investigate the reliability and robustness of the post-hoc explanation techniques LIME and SHAP. They devise a \emph{scaffolding} technique to attack these techniques, allowing them to hide biases of models and give arbitrary explanations.

\cite{saisub2020clusterinterpret} investigate the trade-off between accuracy and interpretability in clustering. They define a measurement for the interpretability of a cluster, by finding the feature value that most of the cluster nodes share. Thus, interpretability is defined as measurement of similarity. With this measurement, the authors show experimentally how to trade-off interpretability and clustering value in a clustering. However, there aren't any human trials to benchmark this definition of interpretability yet.

\cite{hooker2019interpretabilitymethods} detail an automated benchmark for interpretability methods for neural networks using removal of input dimensions and subsequent retraining. While this approach can measure the correlation between explained and actual importance of features, it does not take into account human perception and biases.

\cite{mohseni2018humanevaluation} propose a human-grounded benchmark for evaluating explanations. Human annotators create attention masks that are compared with saliency maps generated by XAI algorithms. A trial showed differences in human and AI explanations, for example a human focus on facial features when recognizing animals, while saliency maps had no focus on specific body parts. In addition, they showed human bias towards different explanation errors.

\cite{schneider2020deceptiveexplanations} investigate trust in explanations. They show that AI outperforms humans at detecting altered, deceptive explanations. 

\cite{10.1145/2939672.2939778} evaluate LIME with both expert and non-expert users, making them choose between different models based on explanation quality and measuring understanding of the underlying model. These experiments show the usefulness of explanations for machine learning-related tasks and show first indications of aspects of explanation quality.

\cite{kaur2020interpretinginterpretability} investigate the use of interpretability by data scientists. A survey showed that data scientists overestimate the quality of XAI methods and misinterpret the explanations provided. While this work does not focus on end users, it highlights the importance of clearly communicating the limits of XAI and investigating how explanations are used and interpreted by humans in real-life scenarios.

\section{VitrAI Prototype for XAI}
\label{sec:prototype}

 \emph{VitrAI} is a web-based service with the goal of demonstrating XAI algorithms in the context of real-life scenarios and evaluating performance and comprehensibility of XAI methods by non-specialists.

The platform consists of two core sections. The first one is the demo section, in which XAI explanations are exhibited in three scenarios. These are described in \ref{subsubsec:Demo section}. The purpose of the demo section is to provide a user-friendly introduction to XAI, requiring no setup or data input. It is intended as an interactive demo for talks, exhibitions and lectures.

The other one is the user-controlled section, where users can choose from several pre-trained machine learning models and provide own data input with subsequent explanations by XAI methods. This section allows for deeper examination and experimentation with XAI methods, for example during user testing.

For each section, different XAI methods are implemented, they are depicted in \ref{subsec:Supported XAI methods}. \emph{VitrAI's} name is a compound of the Latin word vitrum (glass) and Artificial Intelligence.

\subsection{Machine Learning Tasks and Data Types}
\label{subsec:Machine learning and data types}

Across the platform, a mix of general and domain-specific tasks and datasets is encountered. On the one hand, there is unstructured data like images and texts. On the other hand, classical tabular data is present. Each dataset belongs to a machine learning task. Presently, XAI is closely related to supervised learning tasks~\cite{10.1145/3359992.3366639}, hence all considered tasks are supervised in nature.

The following machine learning tasks are considered: supervised text classification, supervised image classification and supervised binary classification of tabular data. For such tasks, many machine learning models have been trained and deployed in the last few years, in science as well as in the industry. Hence, these tasks are suited for the evaluation of XAI methods.

\subsection{Main Sections}
\label{subsec:Main sections}

\subsubsection{Demo Section}
\label{subsubsec:Demo section}
\noindent The foundation of \emph{VitrAI's} demo functionality consists of three real-life scenarios. The first one is called \emph{Public Transport} and is about supervised text classification with five classes. It contains six user complaints about public transport in German language. These complaints are based on true events and revolve around buses leaving too late or too early, unfriendly bus drivers or service issues related to public transport. Each text or complaint is assigned a label, e.g. "ride not on time" or "wrong bus stop", by a machine learning model. The used machine learning algorithm is a Convolutional Neural Network built and trained with spaCy\footnote{\url{https://spacy.io/}} and achieves an accuracy score of 92\%.

The second scenario is called \emph{Car Brands} and is a supervised image classification task. This scenario deals with car images from various angles and distances. The seven images are gathered from the CompCars dataset\footnote{\url{http://mmlab.ie.cuhk.edu.hk/datasets/comp_cars/index.html}} and are assigned one of seven make labels, e.g. "Volkswagen" or "Skoda". The machine learning model is a fine-tuned transfer learning model and achieves an accuracy score of 87\%.

The third setting, \emph{Weather Forecast}, is a supervised binary classification task. It deals with five samples from the "Rain in Australia" dataset\footnote{\url{https://www.kaggle.com/jsphyg/weather-dataset-rattle-package}} that displays 10 years of weather observations from locations across Australia. From a data focused point of view, this data is an example of tabular data, i.e. numerical and categorical features like wind direction, wind speed in the afternoon or atmospheric pressure in the morning of the day before. The machine learning model predicts rainfall for the next day in a binary fashion ("Yes" or "No"). We used a tree-based model trained with scikit-learn\footnote{\url{https://scikit-learn.org/stable/}} that shows an accuracy score of 84\%.

Next to the three scenarios there is a \emph{Dataset Information} functionality that, based on Pandas Profiling\footnote{\url{https://github.com/pandas-profiling/pandas-profiling}}, gathers dataset facts and statistics like variable types, warnings about missing features, distributions of features and correlation matrices. With this built-in functionality, users can get an overview of tabular data before diving deeper by using XAI algorithms. Even with well-functioning XAI methods a sound understanding of the underlying data is obligatory which is why we decided to include the \emph{Dataset Information} functionality.

Note that in the demo section, the training of custom machine learning models is not possible. Instead, several pre-trained models exist for each scenario. These existing models were trained with the goal of achieving a decent accuracy score in order to subsequently obtain reasonable explanation quality. At the same time, the models should not be overly complex, so that an average specialist can build them. For example, in the text scenario the machine learning model was built and trained with spaCy in a straightforward manner and achieves an accuracy score of 92\%. These requirements make for a setting in which XAI methods are used in practice and can therefore be realistically evaluated.

\begin{figure}[h]
 \includegraphics[width=\textwidth]{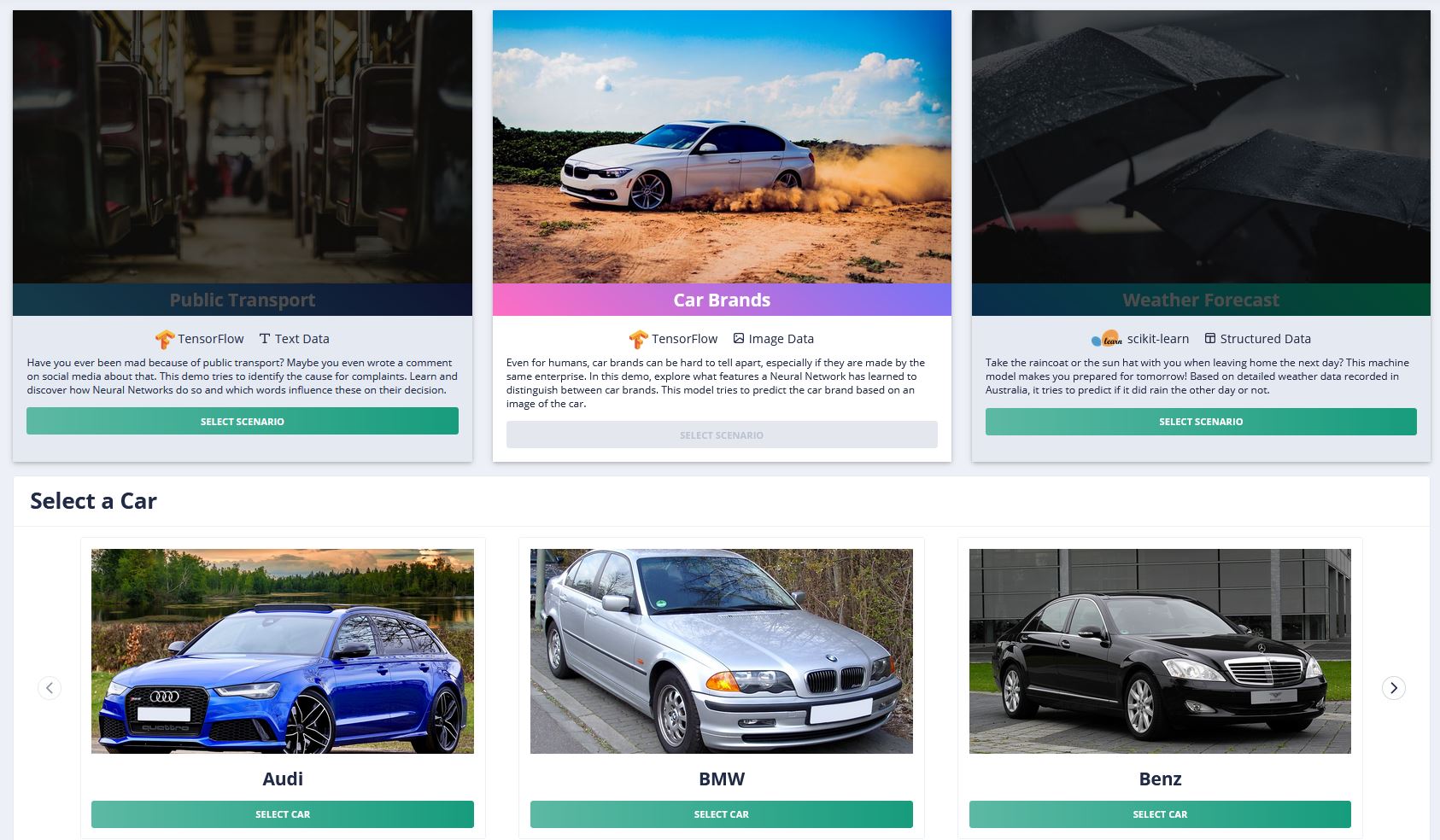}
 \caption{Overview of the demo section with selected \emph{Car Brands} scenario.}
 \label{fig:images/demosection}
\end{figure}

An overview of the demo section is depicted in \autoref{fig:images/demosection}.

\subsubsection{User-controlled Section}
\label{subsubsec:User-controlled section}

\noindent In comparison to the demo section, in the user-controlled section users can create custom machine learning models and predict self-provided samples with subsequent explanations. Regarding training, only text models can be created. For this purpose, an earlier software development project at Fraunhofer IAO is used. Here, a custom text classification model can be trained on a user-defined dataset with common machine learning libraries. 

\begin{figure}[h]
\includegraphics[width=\textwidth]{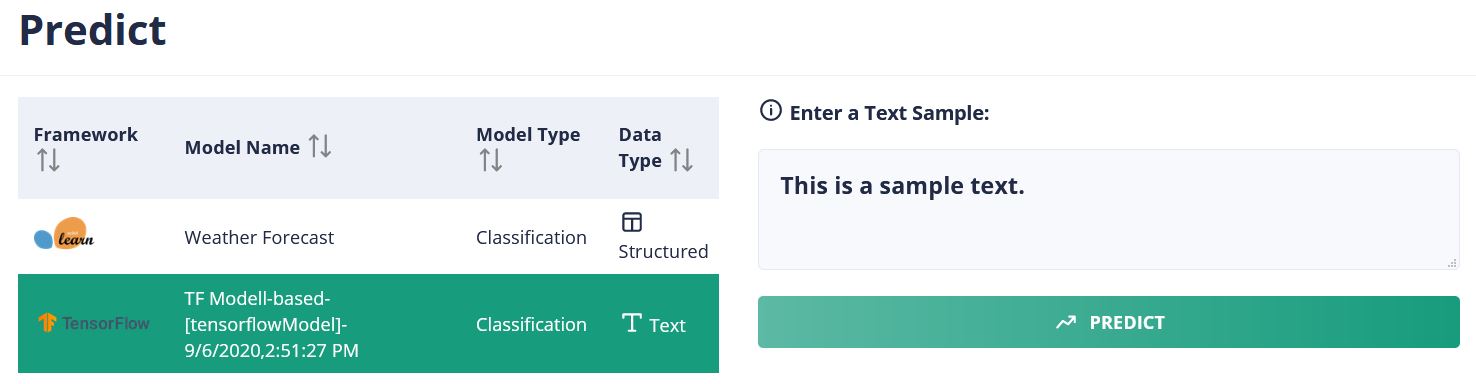}
\caption{Users first choose from a list of pre-trained models and, in the case of a text classification model, can provide a text sample which will be classified and explained later.}
\label{fig:images/predict_text}
\end{figure}

Concerning prediction, users can select pre-trained machine learning models (and thus, a task and data type). In the case of a text classification model, the user can enter a text sample in the user interface (see \autoref{fig:images/predict_text}). Similarly, in the case of an image classification model, a sample image can be uploaded and when choosing a binary classification model, a single-line csv file containing the features as columns should be uploaded.

\subsection{Supported XAI Methods}
\label{subsec:Supported XAI methods}

\noindent In total, the following four XAI methods are implemented: Layer-Wise Relevance Propagation (LRP)~\cite{Bach2015OnPE}, LIME~\cite{10.1145/2939672.2939778}, SHAP~\cite{NIPS2017_7062}, and scikit-learn's permutation importances\footnote{\url{https://scikit-learn.org/stable/modules/permutation_importance.html}}. Each scenario features at least two different XAI approaches. \autoref{tab:usedxaimethods} shows the implemented XAI approaches for each scenario.

\begin{table}
\caption{Considered scenarios with implemented XAI methods}
\label{tab:usedxaimethods}
\begin{center}
\begin{tabular}{r@{\quad}c@{\quad}c@{\quad}c@{\quad}c}
\hline\rule{0pt}{12pt}
& LRP& LIME& SHAP& Permutation Importances\\[2pt]
\hline\rule{0pt}{12pt}
Public Transport & \checkmark& \checkmark& & \\
Car Brands & & \checkmark& \checkmark& \\
Weather Forecast & & \checkmark& \checkmark& \checkmark \\
\hline
\end{tabular}
\end{center}
\end{table}

The used XAI algorithms are well-studied and established in the field of XAI~\cite{Tjoa2020ASO}. With permutation importances,  \emph{VitrAI} does not only support instance-based explanations but also an explanation approach related to whole machine learning models. It is understood that permutation importances provide a rather rudimentary form of model insight, but due to their low complexity and easy usage, they can readily be used by practitioners.

\begin{figure}[h]
 \includegraphics[width=\textwidth]{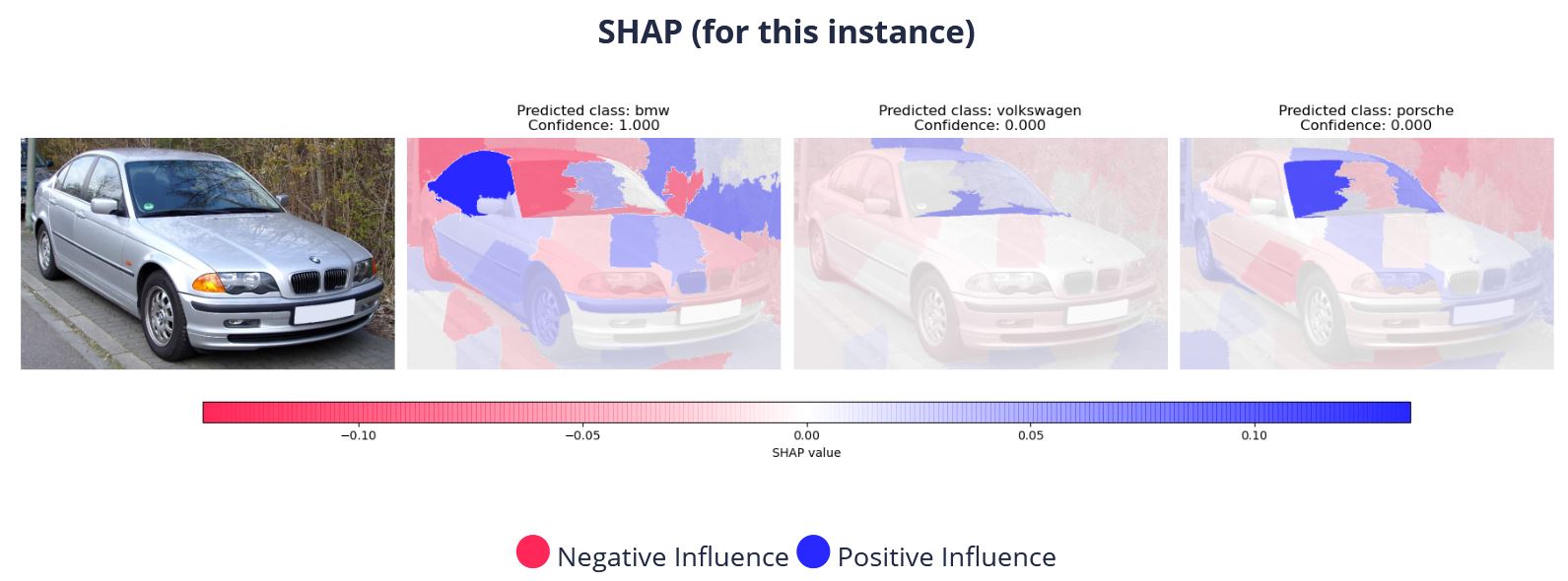}
 \caption{SHAP explanation for the true prediction (nearly 100\% class probability) of the car make \emph{BMW}.}
 \label{fig:images/shap_car}
\end{figure}

\begin{figure}[h!]
	\begin{center}
	\includegraphics[scale=0.4]{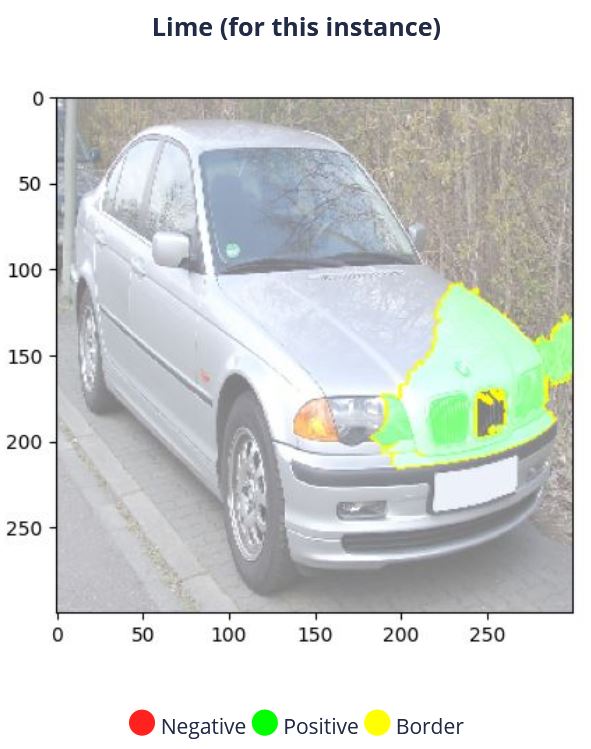}
	\end{center}
	\caption{LIME explanation for the true prediction (nearly 100\% class probability) of the car make \emph{BMW}.}
	\label{fig:images/lime_car}
\end{figure}

\begin{figure}[h!]
	\begin{center}
	\includegraphics[scale=0.4]{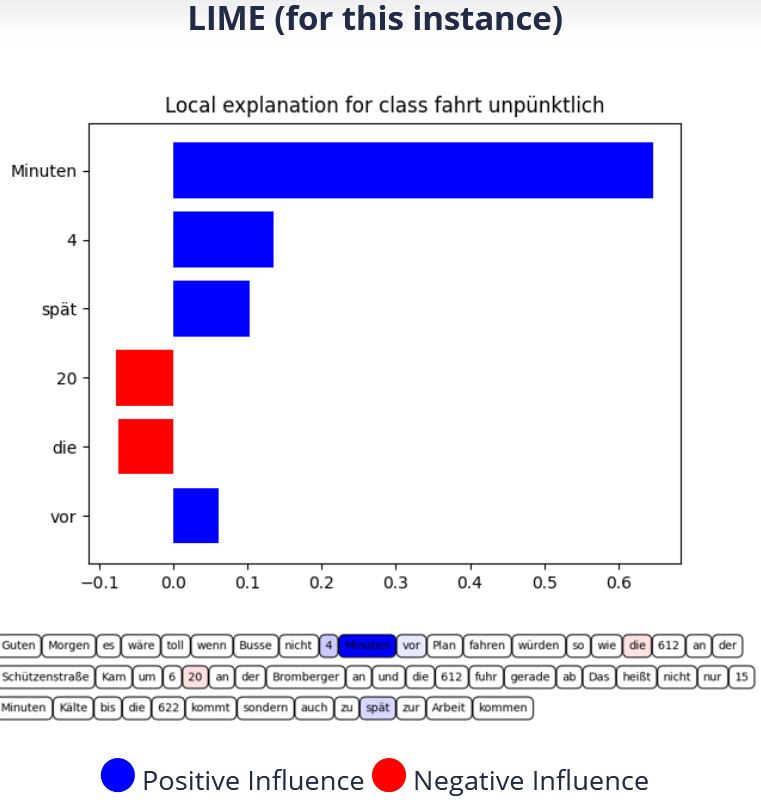}
	\end{center}
	\caption{LIME explanation for the prediction of a complaint type.}
	\label{fig:images/lime_text}
\end{figure}

\begin{figure}[h!]
 \includegraphics[width=\textwidth]{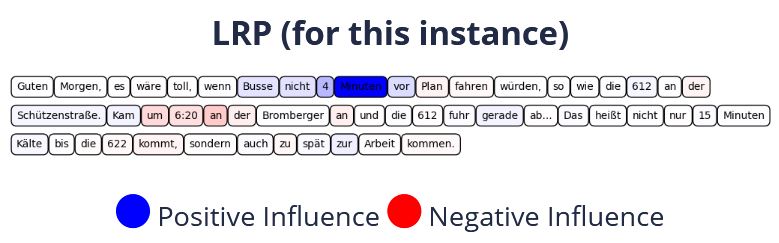}
 \caption{LRP explanation for the prediction of a complaint type.}
 \label{fig:images/lrp_text}
\end{figure}

\autoref{fig:images/shap_car}, \autoref{fig:images/lime_car}, \autoref{fig:images/lime_text} and \autoref{fig:images/lrp_text} show the XAI algorithms used in the Text and Image Classification scenario.

\subsection{Architecture and Technologies Used}

\noindent  \emph{VitrAI} is built according to the microservice pattern. This means that functionally independent components are encapsulated as stand-alone applications in Docker\footnote{\url{https://www.docker.com/}} containers. The containers are orchestrated with Docker Compose. The core functionality of  \emph{VitrAI} is in the modules, which are divided according to the type of data processed. For the frontend, Angular\footnote{\url{https://angular.io/}} is used in order to apply the component libraries Nebular\footnote{\url{https://akveo.github.io/nebular/}} and PrimeNG\footnote{\url{https://www.primefaces.org/primeng/}}. For the backend, technologies used are Django\footnote{\url{https://www.djangoproject.com/}}, Flask\footnote{\url{https://flask.palletsprojects.com/en/1.1.x/}}, CouchDB\footnote{\url{https://couchdb.apache.org/}} and the machine learning libraries TensorFlow\footnote{\url{https://www.tensorflow.org/}}, Keras\footnote{\url{https://keras.io/}}, scikit-learn and spaCy. 

\section{Preliminary Findings}

\noindent To begin with, we experienced challenges regarding uniform implementation and representation of different XAI methods. Although there are frameworks that orchestrate different methods, they either lack XAI approaches, present explanations heterogeneously or show dependency shortcomings (for example, iNNvestigate~\cite{JMLR:v20:18-540} and AI Explainability 360~\cite{aix360-sept-2019} lack support of TensorFlow 2.0/tf.keras models). The last issue also applies to different XAI implementations and overall results in challenging usability (e.g. existing machine learning models may need to be retrained for compatibility reasons). 

We preliminarily evaluated the image and text classification tasks. In the \emph{Car Brands} scenario, LIME highlights background regions and therefore irrelevant regions in four out of seven samples (see \autoref{fig:images/lime_car} for an example). This applies to positive as well as negative influences. On the other hand, in every sample at least parts of the highlighted regions are plausible, e.g. the brand logo or characteristic grilles are marked as positive influences (see part of the grille in \autoref{fig:images/lime_car}). SHAP highlights incomprehensible image regions in every sample (see again \autoref{fig:images/shap_car}). In five out of seven images, at least parts of the proclaimed influences are plausible. In summary, SHAP explanations are incomprehensible more frequently than LIME's, and overall explanation quality is in need of improvement. Also, explanations match human explanations at most partly (among others, humans probably would mark both grilles in \autoref{fig:images/lime_car} as positive influences).

In the \emph{Public Transport} scenario, LRP produced helpful explanations in three out of six cases. A positive example can be seen in \autoref{fig:images/lime_text}, where "minutes", "4" and "late" are assigned positive influences for the sample's association to the class "ride not on time". Less plausible is LRP's proclamation of the word "stop" as a negative influence when dealing with the class "stop missed by bus". LIME shows slightly better explanations in as much as we find four out of six explanations to be acceptable. In for out of six cases, we assess the explanations of both algorithms as similar, while at least partial conformity with human explanation is given in three out of six samples.

Summing up for all three cases, different XAI algorithms show only partial agreement among themselves and match human expectations unsystematically. For the unsatisfactory explanations, it is not clear whether they are due to poor performance of the machine learning model, poor performance of the XAI model, or some other reason.

\section{Conclusion and Outlook}

In this paper we presented \emph{VitrAI}, a platform for demonstrating and evaluating XAI. We have outlined the main modules and XAI approaches used as well as given preliminary evaluation results, showing a need for further improvement and investigation.

In the future, we intend to add more XAI approaches, e.g. CEM Explainer~\cite{10.5555/3326943.3326998}, Lucid\footnote{\url{https://github.com/tensorflow/lucid}} or ProtoDash~\cite{DBLP:conf/icdm/GurumoorthyDCA19}. Since different methods pursue diverse explanation strategies (post-hoc/ante-hoc explanations, local/global direct explanations), for a thorough evaluation a most complete coverage of different approaches is necessary.

More importantly, we will be using  \emph{VitrAI} to further research in the area of explanation quality, as it regards end users. This evaluation shall be conducted by means of aforementioned three scenarios with involvement of humans. The focus will be on the following research questions: 

\begin{enumerate}
\item How do non-specialists assess the comprehensibility of different XAI explanations?
\item What constitutes explanation quality for end users? Can it be measured, compared and quantified?
\item To what extent do XAI explanations conform to human explanations? What are possible reasons for discrepancies?
\item How can explanation quality in XAI be improved for end users?
\end{enumerate}

An experimental evaluation with human involvement promises a better understanding of current practical shortcomings of XAI approaches and can therefore initiate future research leading to increased acceptance of XAI explanations in practice.

\subsubsection*{Acknowledgment}
This work was conducted together with students from the University of Stuttgart.

%
\bibliographystyle{bibtex/spmpsci}
\bibliography{bibliography}
\end{document}